\documentclass{article}

\PassOptionsToPackage{numbers, compress}{natbib}
\usepackage[final]{nips_2018}

\usepackage[utf8]{inputenc} % allow utf-8 input
\usepackage[T1]{fontenc}    % use 8-bit T1 fonts
\usepackage{hyperref}       % hyperlinks
\usepackage{url}            % simple URL typesetting
\usepackage{booktabs}       % professional-quality tables
\usepackage{amsfonts}       % blackboard math symbols
\usepackage{nicefrac}       % compact symbols for 1/2, etc.
\usepackage{microtype}      % microtypography

\usepackage{verbatim}
\usepackage{amsmath}
\usepackage{graphicx}

\title{Learning Individualized Cardiovascular Responses from Large-scale Wearable Sensors Data}

\author{Haraldur T. Hallgr\'imsson\thanks{Work done while interning at Evidation Health.} \\ Department of Computer Science \\ University of California, Santa Barbara \\ \texttt{hth@cs.ucsb.edu}
\And
Filip Jankovic \\ Evidation Health \\ Santa Barbara, CA \\ \texttt{fjankovic@evidation.com}
\And
Tim Althoff \\ Paul G. Allen School of Computer Science \& Engineering \\ University of Washington \\ \texttt{althoff@cs.washington.edu}
\And
Luca Foschini \\ Evidation Health \\ Santa Barbara, CA \\ \texttt{luca@evidation.com}}

\begin{document}

\maketitle

\begin{abstract}
We consider the problem of modeling cardiovascular responses to physical activity and sleep changes captured by wearable sensors in free living conditions. We use an attentional convolutional neural network to learn parsimonious signatures of individual cardiovascular response from data recorded at the minute level resolution over several months on a cohort of 80k people.
We demonstrate internal validity by showing that signatures generated on an individual's 2017 data generalize to predict minute-level heart rate from physical activity and sleep for the same individual in 2018, outperforming several time-series forecasting baselines. We also show external validity demonstrating that signatures outperform plain resting heart rate (RHR) in predicting variables associated with cardiovascular functions, such as age and Body Mass Index (BMI). We believe that the computed cardiovascular signatures have utility in monitoring cardiovascular health over time, including detecting abnormalities and quantifying recovery from acute events. 
\end{abstract}

\section{Introduction}
When engaging in any physical task, the human body responds through a series of integrated changes in function that involves its physiologic systems, including the musculoskeletal, the cardiovascular, and the respiratory systems~\cite{brooks1996physiologic}. Such responses may vary significantly due to environmental factors, yet when elicited in a controlled environment such as a 6-minute walk test carried out in lab settings, they allow inferring individual-specific physiological markers such as Resting Heart Rate (RHR), Maximal Heart Rate, and Maximal Stroke Value. These markers are important in characterizing an individual's health and fitness status. For example, it is well known that cardio-respiratory fitness is inversely associated with all-cause mortality~\cite{doi:10.1001/jamanetworkopen.2018.3605}.

Recently, the advent and widespread adoption of wearable devices and fitness tracking apps~\cite{patel2017using} has enabled continuous, unobtrusive tracking of an individual behavior and physiological signals such as heart rate, physical activity, and sleep over time, with time resolution down to the minute-level and below. This has enabled population-scale physiological sensing~\cite{althoff2017harnessing}.

In this work, we move beyond population-level aggregated sensing and set out to learn \emph{individual} characteristics of cardiovascular responses by observing the relationship between behaviors such as sleep and physical activity and their associated changes in heart rate during the individuals everyday life.
In absence of the controlled lab settings usually described in the physiology literature~\cite{brooks1996physiologic}, we hypothesize that prolonged observation periods increase the likelihood of a behavior mimicking an in-lab test to spontaneously occur. For example, a brisk walk to the train station may be a good approximation of a 6-minute walk test. For this reason, we make use of attentioned models to pick up on such ``natural experiments'' that collectively contribute to shaping the envelope of an individual's cardiovascular response. In an analogy with control theory, we set out to learn the cardiovascular \emph{transfer function} of an individual to capture the cardiac output for each possible (behavioral) input.

Though previous studies have leveraged representation learning to extract health-related features from wearable sensor data~\cite{quisel2017collecting, deepheart}, our work is unique in terms of dataset size ($2.6\times 10^{9}$ minutes of sensor measurements considered from 80k users over a span of one year), outputs (parsimonious individualized cardiovascular signatures output by attentioned convolutional autoencoders), and validation methods. We believe that accurately capturing cardiovascular response enables screening for abnormalities and detecting physiological changes over time unobtrusively in free living conditions.

\section{Data}

\begin{figure}
  \centering
  \includegraphics[width=1.0\textwidth]{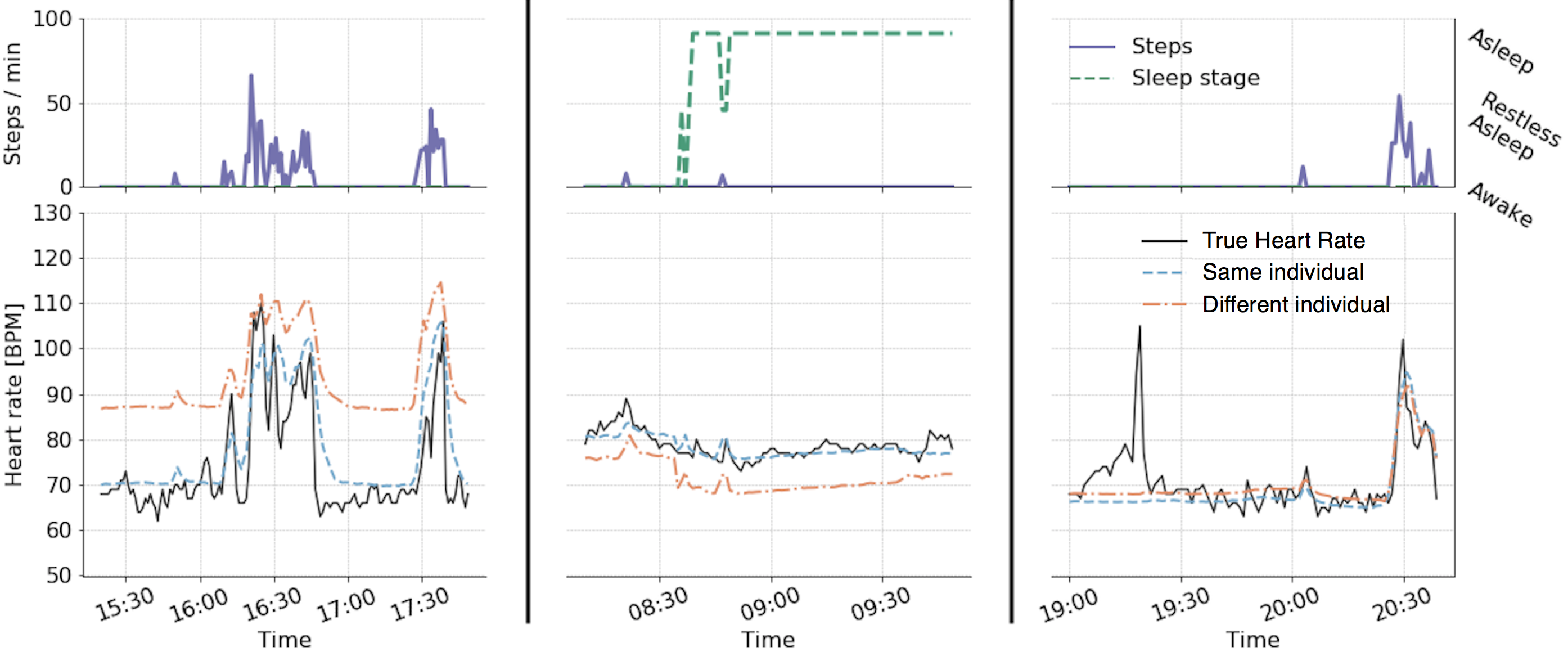}
  \caption{Example physical activity and sleep (upper row) and heart rate (bottom row) sensor data from three individuals, demonstrating how heart rate responds to onset of exercise (left column) and sleep (middle column). Changes in heart rate do not always occur due to physical activity (right column), with onset of anxiety or stress being potential unmeasured confounders. As expected, applying a signature from a different person (demonstrated in orange) results in increased reconstruction error.}
  \label{fig:sample_data}
\end{figure}

We select a cohort of 80,137 members of Achievement, a commercial reward platform. To be included in this study, users must have authorized sharing with Achievement of dense minute-level steps/sleep/heart rate activity logs from commercial activity trackers, such as Fitbit or Apple Watch. 
Following~\citep{migueles2017accelerometer}, to be included in the cohort a member must have at least 10 days worth of physical activity logs, with no more than 4 hours of unreported data per day, for one or both of the collection windows of January 2017 or 2018. %On average, each member had 32,750 minutes of reported data per month, with half the members reporting between 26,488 to 40,537 minutes per month. 82.8\% of this cohort is female, with a median age and BMI of 31 and 28.3, respectively.
Half of the members reported between 26,488 to 40,537 minutes per month, averaging 32,750 minutes. 82.8\% of this cohort is female, with a median age and BMI of 31 and 28.3, respectively.
All members with reported data in both of the two months were assigned to the validation set (N=25,406). The remaining individuals were randomly assigned to either the training (N=43,784) or tuning sets (N=10,947).

%In the health survey, each member indicated a binary variable of being diagnosed with one or more of several health conditions. We focus on those conditions which we believe could be associated with the heart rate response to physical activity: type-2 diabetes, cardiovascular disease, stroke, obstructive sleep apnea, ...

The data from the activity trackers are minute-level measurements of a person's total step count and average heart rate, and if the wearer is asleep or restless asleep; see Figure~\ref{fig:sample_data} for sample data from three individuals. We scaled these measurements to the range $(0, 1)$ to speed up model training~\cite{lecun2012efficient}: the heart rate per-person is whitened to zero mean and unit standard deviation, and the step count values are log-transformed to handle the large spread of values as: $\text{steps}' = \log\left(\text{steps} + 1\right)/5$. The two sleep stages are encoded as separate binary channels. Missing data is imputed as mean heart rate of activity at awake, and no other data cleaning is performed.

\section{Cardiovascular Signature Network}

%We aim to learn a signature of a person's cardiovascular response to their everyday activity that can be used to predict that person's heart rate response. %Such a signature should accurately capture how their heart responds to extreme events such as the onset and offset of exercise. 
To learn a personalized cardiovascular response function, we consider a heart rate autoencoder~\cite{bengio2013representation} that is conditioned on the physical activity and sleep stages.
%two-stage model that acts as an autoencoder~\cite{bengio2013representation} on the heart rate signal.
The signature-encoder learns a signature of a person based on how their heart rate responds to physical activity, while the signature-decoder uses a learned signature to predict a person's heart rate based on their physical activity. %Both models need to consider sufficiently long time-dependencies at multiple timescales, and should focus on extreme and rare events that stress a persons cardiovascular response.

%\subsection{Model architecture}

\begin{figure}
  \centering
  \includegraphics[width=0.65\textwidth]{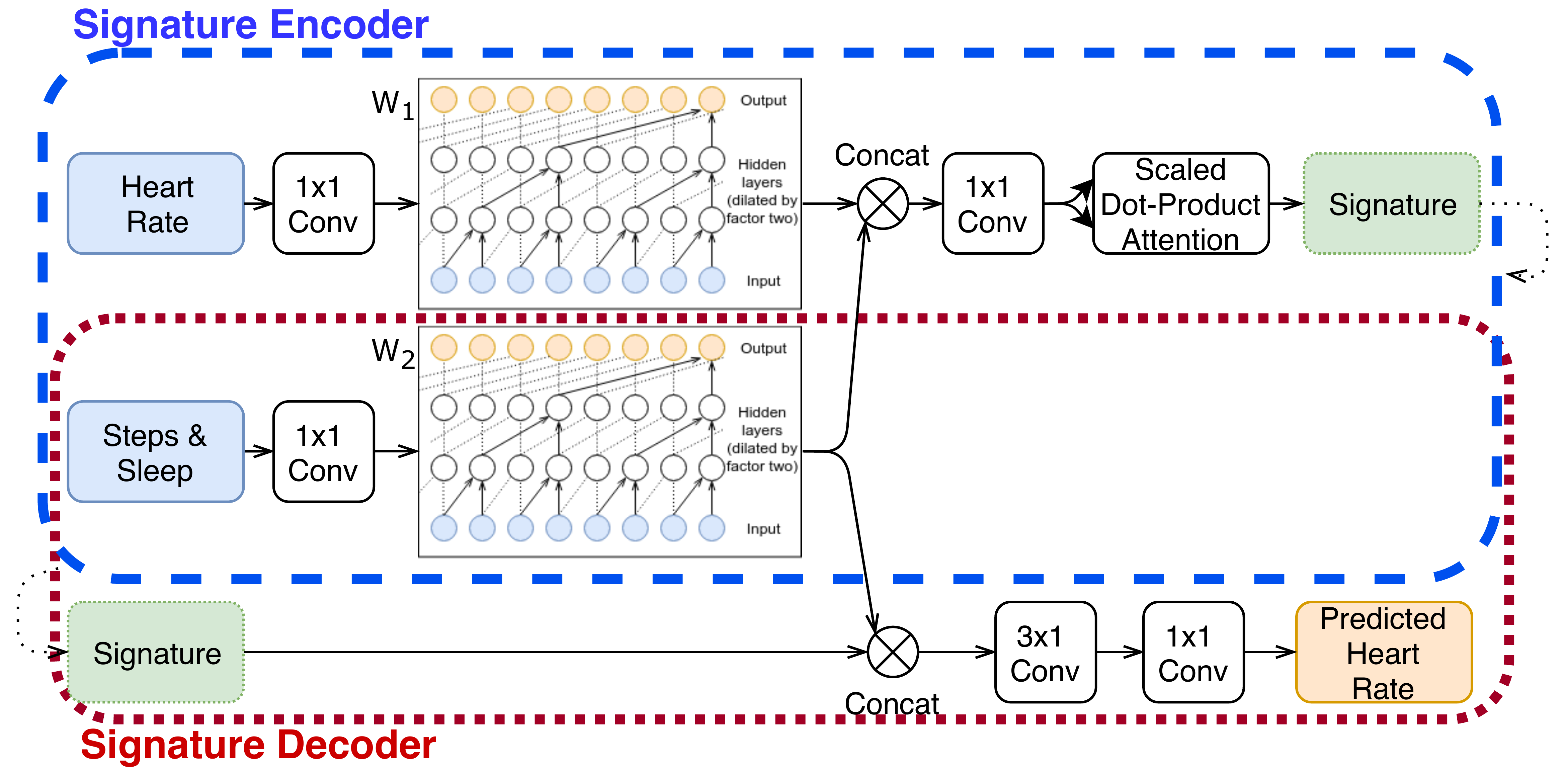}
  %\fbox{\rule[-.5cm]{0cm}{4cm} \rule[-.5cm]{4cm}{0cm}}
  \caption{Diagram of proposed model architecture. The signature encoder predicts a cardiovascular signature from measured sensor data (top dashed box), and the signature decoder uses that same signature as well as physical activity data to predict the heart rate (bottom dotted box).}
  \label{fig:model_arch}
\end{figure}

%In modeling a cardiovascular signature, several attributes of the problem should be kept in mind. Chief among them is that the heart rate response is a non-linear, time dependent system. The model should also be able to consider sufficiently long time-dependencies and at multiple timescales to capture the entirety of the heart rate recovery curve. Finally, in learning a personalized transfer function, most minutes of physical activity are not informative; instead, we expect extreme and rare events that stress a persons cardiovascular response to be most revealing. 

\textbf{Encoder:} 
The encoder model, as seen in Figure~\ref{fig:model_arch}, learns a fixed-size signature from an arbitrarily length time-series. It consists of two WaveNet~\cite{van2016wavenet} convolutional neural network (CNN) blocks, $W_1$ and $W_2$, composed of seven dilated causal convolutional layers with residual connections and allow for modeling long-range temporal dependencies of up to 128 minutes, with 32 and 16 filters per layer, respectively. As opposed to recurrent layers, convolutions are typically faster to train especially when applied to very large sequences such as considered here. The encoder considers the physical activity channels and the heart rate signal separately in $W_1$ and $W_2$, which allows the encoder to jointly learn a latent physical activity representation with the decoder by sharing the weights of $W_2$. The outputs of $W_1$ and $W_2$ are concatenated together and a scaled dot-product attention mechanism~\cite{vaswani2017attention} is applied to predict the cardiovascular signature with queries and keys of dimension $d_k = d_v = 8$ while the dimensionality of the values, $d_v$, is equal that of the signature size. Three separate convolutional layers of filter width 1 are applied to re-size the tensors appropriately.

\textbf{Decoder:} 
The decoder model consists of a single WaveNet block $W_2$, whose weights are tied to that of the encoder's, followed by two temporal convolutional layers. The output of $W_2$ at every time step is concatenated with a signature vector, and two temporal convolutional layers are then trained to predict the corresponding heart rate signal. The number of parameters unique to the decoder are kept to a minimum to force the signature to be as informative as possible.

\textbf{Training:} The two models are learned end to end by minimizing the average $L_2$ norm of the error in predicting heart rate, using the Adam optimizer~\cite{kingma2014adam} with default parameters ($\alpha=0.001$, $\beta_1=0.9$, $\beta_2=0.999$, $\epsilon=10^{-7}$). Missing data is imputed as mean heart rate of no activity at awake, though no loss is propagated corresponding to these periods. The models are implemented in Keras~\cite{chollet2015keras} with a TensorFlow~\cite{abadi2016tensorflow} backend. All hidden layers include ReLU activation functions~\cite{nair2010rectified}, with the exception of the WaveNet blocks, which use gated activation units~\cite{van2016wavenet}, and the output, which has no non-linearity. Training was done on mini-batches of size 16, for up to 30 epochs with an early stopping criteria if validation error was not observed to improve for five epochs.

\section{Experimental results}
\textbf{Baseline models:}
We consider three baselines to compare our model to. The simplest predicts a persons mean heart rate at awake or asleep. The second uses XGBoost~\cite{chen2016xgboost} with default parameters, trained on a single person to predict their heart rate based on the previous 120 minutes of physical activity. The third uses XGBoost again, but this time trained on a population of people rather than at the individual level. The performance of the baseline models can be seen in Table~\ref{tab:results}. Both XGBoost models are trained on the January 2017 activity window and validated on January 2018.

\begin{table}
  \caption{Experimental results. The trained proposed model was validated on the 2017 data, and also using 2017 signatures applied to 2018 data. While varying signature sizes (results shown in middle column) the full training set was used, and when varying training set size (results shown in right column) a size-32 signature was used. For comparison, we include the performance of the baseline models (in left column) trained on 2017 data and validated on 2018 data.}
  \label{tab:results}
  \centering
  \begin{tabular}{lr|rrl|rrl}
    \toprule
    &&\multicolumn{3}{c|}{\textbf{Varying signature size}} & \multicolumn{3}{c}{\textbf{Varying training set size}} \\
    \cmidrule(r){3-8}
    \textbf{Baseline model} & \begin{tabular}{@{}c@{}}\textbf{Validation} \\ \textbf{error}\end{tabular} & \textbf{Size} & \multicolumn{2}{c|}{\begin{tabular}{@{}c@{}}\textbf{Validation error} \\\textbf{2017 / 2018}\end{tabular}} & \textbf{\# people} & \multicolumn{2}{c}{\begin{tabular}{@{}c@{}}\textbf{Validation error} \\ \textbf{2017 / 2018}\end{tabular}} \\
    \midrule
    Awake/Sleep Mean & 0.755  & 4 & 0.295 & 0.385 & 500 & 0.319 & 0.400 \\
    Individual XGBoost & 0.445 & 8 & 0.291 & 0.385 & 2,000 & 0.306 & 0.391 \\
    Population XGBoost & 0.539 & 16 & 0.283 & 0.394 & 5,000 & 0.298 & \textbf{0.383} \\
    && 32 & 0.279 & 0.393 & 20,000 & 0.285 & 0.387 \\
    && 64 & 0.288 & \textbf{0.384} & 43,784 & \textbf{0.279} & 0.393\\
    && 128 & \textbf{0.278} & 0.395 &&& \\
    \bottomrule
  \end{tabular}
\end{table}

\textbf{Sensitivity analysis on signature size:}
 We trained the proposed model with signature sizes of $\{4, 8, 16, 32, 64, 128\}$. As seen in Table~\ref{tab:results}, we observe that our model is robust to varying sizes of the cardiovascular signatures, with a decrease in validation error that levels off after a size of 16.

\textbf{Effect of training set size:}
Our model leverages a population to learn a single persons cardiovascular transfer function. To understand the effect of the population on the model, we vary the training set size as fractions of the total (1\%, 5\%, 10\%, 50\%, 100\%) and observe how well our model performs. As seen in Table~\ref{tab:results}, we observe a steady decrease in validation error as the training data is increased, culminating in a 14\% better performance with a full dataset as opposed to only 1\% of it.

\textbf{Signature consistency:}
To assess test-retest reliability, a measure of \emph{internal validity}, we consider how well a cardiovascular signature can be used to predict a persons heart rate from their physical activity a year later. For each individual in the validation set, we learn a signature from their signal measurements during January 2017 and apply that signature to predict their heart rate during January 2018. As compared to using a different persons signature, a person's own signature is significantly better at predicting their heart rate (Wilcoxon signed-rank test, $V=2.6\times 10^7$, $p<10^{-16}$), with a median of 60\% greater mean-square error when using another person's, randomly selected.

\textbf{Predicting health conditions using signatures:}
To assess the \emph{external validity} of the signatures, we tested whether they are associated with factors affecting cardiovascular response, such as age and body mass index (BMI). We used an XGBoost model~\cite{chen2016xgboost} trained on the learned size-32 validation signatures to predict if an individual is above/below median age of the cohort (31 years) with an AUC of 70.1\% when trained on a random 70/30 split of the validation set. Predicting if a person is obese (BMI $\ge 30$) from solely their signature achieves an AUC of 69.7\%. Predicting the same outcomes using only an individual's resting heart rate results in significantly worse accuracy, with AUCs of 60.6\% and 54.1\%, respectively, demonstrating that signatures carry richer information about the relationship between physical activity and heart rate than the single RHR marker.
\section{Discussion}

It is informative to consider when a cardiovascular signature would \emph{not} well predict a person's heart rate. Assuming the measuring conditions of the wearable device stay the same, this may happen when a person's cardiovascular response is hard to learn (e.g., short observation period, high missingness, or erratic behavior), when it changes (e.g., improvement/degradation of fitness), or when there are factors affecting HR that go beyond sleep and physical activity (e.g., stress endured during an interview, after taking medication, or having a meal). An example of where our model fails can be seen in the right-most column of Figure~\ref{fig:sample_data}.

In future work we plan to explore the motifs surfaced by the attention component of the network, and study how they are related to health outcomes.
From a methods perspective, future extensions will consider variational autoencoders to better condition the latent space of cardiovascular signatures as well as further hyper-parameter and architecture optimization. 
%An interesting follow-up question would be to learn a cardiovascular signature solely from physical activity data. Given a long enough observation period including daily routines, exercise, and other extreme events, can we predict a person's cardiovascular transfer function? This might be impossible in the case of latent variables such as health conditions.

% Reference text size set to small, which is permitted by the NIPS submission guidelines.
{\small
\bibliography{ref}}
\bibliographystyle{plain}

\end{document}